\documentclass[a4paper,11pt]{article}

\usepackage{graphicx}  %%% for including graphics
\usepackage{url}       %%% for including URLs
\usepackage{times}
\usepackage{natbib}
\usepackage[margin=25mm]{geometry}

\usepackage{wrapfig}

\usepackage{linguex}

\usepackage{titlesec}

\titlespacing*{\section}
{0pt}{.35\baselineskip}{.3\baselineskip}
\titlespacing*{\subsection}
{0pt}{.28\baselineskip}{.25\baselineskip}
\titlespacing*{\subsubsection}
{0pt}{.2\baselineskip}{.2\baselineskip}

\newcommand{\visgen}{Visual Genome}
\newcommand{\coco}{COCO}
\newcommand{\referit}{ReferIt}
\newcommand{\refcoco}{RefCOCO}
\newcommand{\grex}{GoogleREX}
\newcommand{\flickr}{Flickr30kEntities}

\usepackage{color}

\title{Natural Language Semantics With Pictures: Some Language \& Vision Datasets and Potential Uses for Computational Semantics}
\date{}

\author{David Schlangen\\
       Department of Linguistics, University of Potsdam, Germany\footnote{Work done while author was at Bielefeld University.}\\
       {\normalsize\texttt{david.schlangen@uni-potsdam.de}}
}

\begin{document}
\maketitle
\thispagestyle{empty}
\pagestyle{empty}

\begin{abstract}

  Propelling, and propelled by, the ``deep learning revolution'', recent years have seen the introduction of ever larger corpora of images annotated with natural language expressions.
  % that are meant to capture some aspects of their meaning.
  We survey some of these corpora, taking a perspective that reverses the usual directionality, as it were, by viewing the \emph{images} as semantic annotation of the natural language expressions. We discuss datasets that can be derived from the corpora, and tasks of potential interest for computational semanticists that can be defined on those. In this, we make use of relations provided by the corpora (namely, the link between expression and image, and that between two expressions linked to the same image) and relations that we can add (similarity relations between expressions, or between images). 
Specifically, we show that in this way we can create data that can be used to learn and evaluate lexical and compositional grounded semantics, and we show that the ``linked to same image'' relation tracks a semantic implicature relation that is recognisable to annotators even in the absence of the linking image as evidence. Finally, as an example of possible benefits of this approach, we show that an exemplar-model-based approach to implicature beats a (simple) distributional space-based one on some derived datasets, while lending itself to \emph{explainability}.
\end{abstract}

\section{Introduction}

In model-theoretic formal semantics, the central semantic notion ``truth'' is explicated as a relation between a sentence and a mathematical structure, its \emph{model}. Semantics textbooks are surprisingly evasive about what exactly this structure is meant to be, other than hinting at that it in some way represents the general ``situation'', or ``world'', that the sentence is taken to be talking about. In any case, what the model as a mathematical structure does is to provide a collection of \emph{individuals} about which the sentence could be talking, and an \emph{interpretation} of the non-logical lexical items occurring in the sentence, in terms of sets of individuals (or tuples of individuals). The collection of individuals is typically called the \emph{domain} $D$, and the set of interpretations $I$, so that a model $M = \langle D, I\rangle$.

It is this intended relation with the world that allows us to see an analogy between these structures and photographic images. A photograph is a frozen moment in time, a representation of how the world was (or looked like) at a certain moment, at a certain place and from a certain perspective. And just as a sentence in formal semantics is evaluated relative to a model, a sentence describing a situation can be seen as true \emph{relative to an image} --- if (and only if) the image \emph{depicts} a situation of the described type. Hence, in a slight reversal of our usual way of talking, we can say that a given \emph{image} does (or does not) make a given sentence true (instead of saying that the sentence is a true description of the image), and we can see the image as a model of the sentence.\footnote{%
  There are interesting subtleties here. In our everyday language, we are quite good at ignoring the image layer, and say things like ``the woman is using a computer'', instead of ``the image shows a woman using a computer'', or ``this is a computer'', instead of ``this is an image of a computer''. This also seems to carry over to tense, where we can say ``is using'', instead of ``was using at the time when the picture was taken''. There are however contexts in which talk about the image \emph{as} image is relevant, and this can happen in large corpora such as discussed here. So this is something to keep in mind.}

What does this sleight of hand buy us? A very large amount of data to play with! The field of computer vision has as one of its central aims to find meaning in pixels -- see e.g., \cite{davies:cv}, \cite{marr:vision} -- and a convenient way of representing meaning is with natural language. It is also a field that has been data-driven for a long time, and so there is a large number of data sets available that in some way pair images with natural language expressions.\footnote{%
  See for example the (incomplete) lists at \url{http://www.cvpapers.com/datasets.html} and \url{https://riemenschneider.hayko.at/vision/dataset/}.}
Recent years have specifically seen the creation of large scale corpora where images are paired with ever more detailed language (e.g., single sentence or even full paragraph captions describing the image content; facts about the image spread over question and answer; detailed descriptions of parts of the image in terms of agents and patients; see references below).\footnote{%
  While there are by now some non-English or even multi-lingual corpora, the majority provide \emph{English} language annotations, including all of those that we discuss here.}
  Given the understanding that all these expressions are meant to ``fit'' to the images that they are paired with, and using the slight conceptual inversion of treating the images as ``truthmakers'' \citep{Fine2017} for the sentences, this gives us an unprecedentedly large set of language expressions that are ``semantically annotated.''\footnote{%
  The corpora we discuss here provide almost 8 million distinct natural language expressions (with many more that can be derived from them). In comparison, the largest ``classical'' semantics resource, the Groningen Meaning Bank \citep{Bos2017GMB}, provides some 10,000 annotated sentences, and the Parallel Meaning Bank \citep{abzianidze-EtAl:2017:EACLshort} another 15,000. There is no competition here, though: the Meaning Bank annotations are obviously much deeper and much more detailed; the proposal in this paper is to view the image corpora discussed here as complementary.
}$^,$\footnote{%
  The relation between images and models is implicit in
  \citep{youngetal:flickr30k}, from where we took inspiration, but not further developed there in the way that we are attempting here. \cite{HuerliBos:spatrel} make an explicit connection between image and models, but only look at denotations; as do \cite{schlaetal:imagewac}.}
As we will show, this gives us material to learn about the lexical and compositional semantics that underlies the use of the expressions.

% These ``annotations'' are admittedly -- for linguists -- of an unusual type, as there are ``analogue'', as it were, but this arguably only makes it more interesting. (This is, after all, how the world comes to us in the most part: analogue, rather than packaged in symbolic representations.) **TODO: Add size comparison with meaning bank.**

\vspace*{.5\baselineskip}\noindent
The \textbf{contributions of this paper} are as follows: 1.) To make explicit a perspective that so far has been taken only implictly in the literature, which is to view images as \emph{models} of natural language expressions; 2.)\ to show by example that taking this perspective opens up interesting data sets for computational semantics questions; 3.)\ specifically, to look at how grounded interpretation functions could be learned from and tested on this data, and; 4.)\ how data can be derived that expresses various implicature relationships; and 5.)\ to show how exemplar-based model building can be used to predict some of those relations. Our code for working with the corpora mentioned here (and some others) is available at \url{http://purl.com/cl-potsdam/sempix}.
% By the time of the conference, we will also publicly release the code with which we converted the corpora into a common format, and code for creating the datasets described here.

% The survey part should be of particular interest to readers who have not yet worked with language \& vision corpora; the concrete proposals for tasks 
% curious about what data is available and how it could be used; interspersed are concrete proposals for novel tasks that could be done with this data. 

\section{Background}
\label{sec:back}

\subsection{The Approach: Learning Semantics From Relations in Corpora}
\label{sec:rel}

\begin{wrapfigure}[11]{r}{0.25\linewidth}
  \begin{center}
\vspace*{-2.3\baselineskip}
\includegraphics[width=0.25\columnwidth]{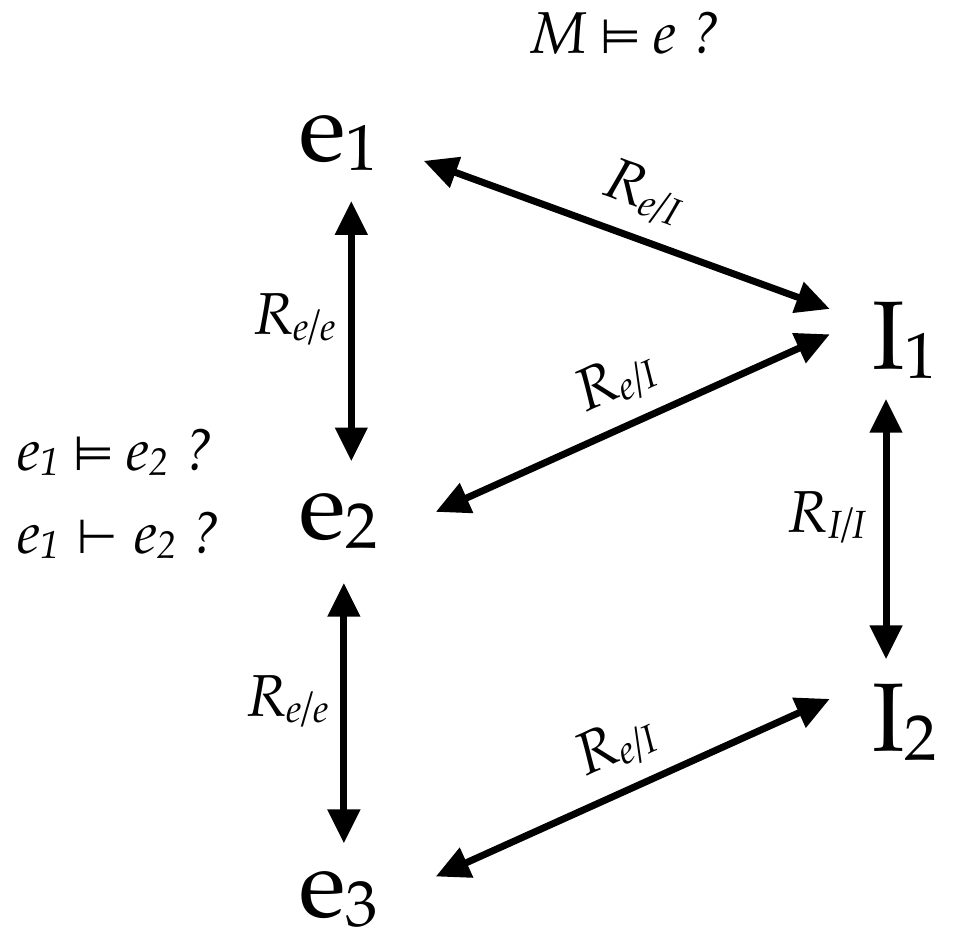}%}
\end{center}
\vspace*{-1.6\baselineskip}
  \caption{Relations in corpora \& to be derived}
  \label{fig:relations}
\end{wrapfigure}

Our general approach will be to look at relations that are expressed in the data or can be added using computational methods, and then to ask what these can tell us about \emph{semantic} relations like truth and entailment, and in turn what these tell us about the meanings of expressions.
Figure~\ref{fig:relations} illustrates the idea. The corpora provide us with an ``annotates'' relation between images and expressions; in the Figure, holding between $I_1$ and $e_1$ and  $e_2$, and $I_2$ and $e_3$, where the expressions for example could be captions. Implicitly, there is also an ``annotates same image'' relation that holds between expressions; here, $e_1$ and $e_2$, as alternative captions of the same image. Standard natural language processing and computer vision techniques (see below) allow us to compute similarity relations between pairs of images (e.g., $I_1$ and $I_2$) and between pairs of expressions ($e_1$, $e_3$). The question then is whether these relations can tell us something about \emph{satisfaction / denotation} ($\models, [\![\cdot]\!]$) and \emph{entailment} ($\models, \vdash$).
%Formally, we can write the sets of relations of the different types as follows: $\mathcal{R}_{e/I} = \{ \mathit{annotates}, [\![\cdot]\!]\}$ (relations between expressions and images), $\mathcal{R}_{e/e} = \{ \mathit{annotates\_same\_I}, sim_{e/e}, \models, \vdash\}$ (relations between expressions), and $\mathcal{R}_{I/I} = \{ sim_{I/I}\}$ (between images).

%This is the question that we explore here; but first we briefly give some details about the corpora that we make use of.

\subsection{Corpora Used Here}
\label{sec:corpora}

We make use of data from the following corpora:

% $\bullet$ \textbf{IAPR TC-12 / ReferIt:} IAPR TC-12 is a collection of ``20,000
%   still natural images taken from locations around the world and
%   comprising an assorted cross-section of still natural images''
%   \citep{Grubinger2006}. This dataset was later augmented by
%   \cite{Escalante2010} with segmentation masks identifying objects in
%   the images (an average of 5 objects per image). It was further
%   augmented by \cite{Kazemzadeh2014} with expressions referring to
%   objects in the images.

$\bullet$  \textbf{MSCOCO / RefCoco / GoogleREX:} The ``Microsoft Common Objects
  in Context (COCO)'' collection \citep{mscoco} contains over 300k
  images with object segmentations (of objects from 80 pre-specified
  categories), object labels, and nearly 400,000 image captions. It was augmented with 280,000 referring expressions by \cite{yueatal:refcoco}, using the ReferitGame where one player needs to get another to identify a predetermined object in the image, with the players getting feedback on their success.
  % :  ``In this two-player game, the first player is shown an
  % image with a segmented target object and asked to write a natural
  % language expression refer- ring to the target object. The second
  % player is shown only the image and the referring expression and asked
  % to click on the corresponding object. If the players do their job
  % correctly, they receive points and swap roles. If not, they are
  % presented with a new object and image for description.''
  \cite{Maoetal:cocorefexp_Final} also provide expressions for COCO
  objects, but collected monologically with the instructions to provide
  an expression that uniquely describes the target object.

$\bullet$  \textbf{Flickr30k / Flickr30kEntities:} Flickr30k
  \citep{youngetal:flickr30k} is a collection of 30,000 images from a
  public image website which were augmented with 160,000 captions;
  \cite{flickr30kent} annotated these captions with positions of
  the objects in the images that they mention (Flickr30kEntities).

$\bullet$  \textbf{Visual Genome:} This dataset by \cite{krishnavisualgenome}
  combines images from COCO and another data set (yielding around 100k
  images), and augments them with 2 million ``region descriptions'', which are 
  statements true about a part of an image, and resolved for the
  entities mentioned and their relations. These descriptions are parsed
  into object names and attributes, and normalised by reference to the
  WordNet ontology \citep{fellbaum:wordnet}.
  \cite{krauseetal:visgenparas} added 20,000 image description paragraphs
  (i.e., extended, multi-sentence captions) for some of the images.

All these data sets give us images paired with natural language
expressions; in most of them, the relation between image and expression
is annotated more fine-grainedly by linking regions within the image to
(parts of) the expressions.\footnote{%
  This makes working with the images easier, as it allows us
to assume that the task of \emph{object recognition} (detecting
contiguous regions of pixels that belong to the same object) has already
been successfully performed. This is not a strict requirement for
working with images these days, however, as high-performing models are
available that do this job \citep{yolov3}, \citep{maskrcnn}, but these
still add noise from which one might want to abstract for the purposes
discussed here.}
Also, some corpora provide an additional layer that could be seen as
corresponding to the logical form of the expression, for example by
normalising nouns to a resource like WordNet \citep{fellbaum:wordnet} or by
annotating the predicate / argument structure.

\section{Expressions and Denotations}
\label{sec:ei}

\subsection{Images as Semantic Models: An Example}
\label{sec:model}

\begin{wrapfigure}[13]{r}{0.5\linewidth}
  \begin{center}
\vspace*{-2\baselineskip}
\includegraphics[width=0.5\columnwidth]{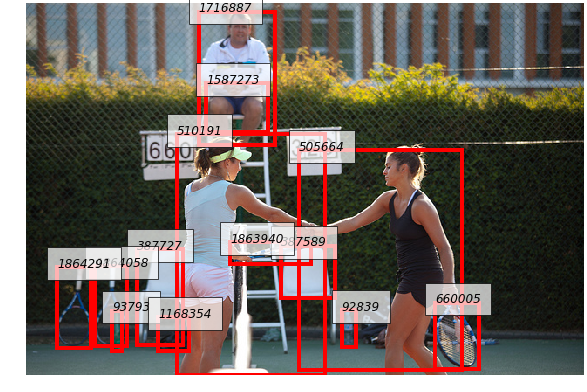}%}
\end{center}
\vspace*{-1.6\baselineskip}
  \caption{A Segmented Image from COCO}
  \label{fig:immod}
\end{wrapfigure}

Above, we have introduced our
% We've made an
analogy between semantic models and images.
% We can make this clearer with an example.
An example shall make it clearer.
Figure~\ref{fig:immod} shows an image (from COCO) with \emph{object segmentations} (rectangular patches indicating the position of an object in the image) and identifiers, as provided by the corpus. We can directly treat this as the \emph{domain} provided by the model, so that here
$D = \{o_{92839}, o_{93793}, o_{387589},$ $ o_{387727}, o_{505664}, o_{510191}, o_{660005}, $ $o_{1168354}, o_{1587273},$ $ o_{1716887}, o_{1863940}, o_{1864058}, o_{1864291}\}$.

The corpus also provides natural language annotations for these objects, for example ``the woman in white'' and ``the woman in black'' (for $o_{505664}, o_{510191}$, respectively). We can use this to ``reverse engineer'' the interpretation functions covering these words, and in particular derive that
$I(woman) \subseteq \{o_{505664}, o_{510191}\}$. If we make an additional \emph{exhaustivity} assumption over the set of annotations, we can strengthen this to $I(woman) = \{o_{505664}, o_{510191}\}$; that is, make the assumption that these are the \emph{only} objects (in this image / the set of segmented objects from that image) to which this term can be applied. We will need to make this assumption when we want to generate \emph{negative instances} used in machine learning, but need to keep in mind that in general, this assumption is unwarranted, as exhaustivity was not a goal when creating the corpora.

\begin{wrapfigure}[5]{r}{0.2\linewidth}
  \begin{center}
    \vspace*{-2.5\baselineskip}
    $o_{505664} =$\\
    \includegraphics[width=0.1\columnwidth]{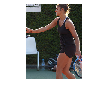}%}
\end{center}
\vspace*{-2\baselineskip}
\caption{Invidual}
  \label{fig:woman}
\end{wrapfigure}
Continuing with the discussion, we can think more about what this view on the corpora offers for doing semantics. Our domain
$D$ is now populated not just with identifiers or symbols from a vocabulary, but rather with objects that have an internal structure. In the example above we were able simply to read off the interpretation function for the word from the annotation. But we can try to use instances like this to \emph{generalise} this function. That is, we can try to turn \(I\) into a ``constructive'' function that not just records a fact (``object $o$ is in the denotation of predicate $\phi$''), but rather produces a \emph{judgement}, given a (structured, visually represented) object;
we may write \(I_{woman}(D)\) to make this explicit.\footnote{%
  This perspective has previously been taken by \cite{schlaetal:imagewac} and developed for simple expressions; the present section builds on that work.
}
% To make this more explicit, we change the notation slightly, and
% make the domain an explicit argument of the function: \(I_{woman}(D)\).
% Seen this way, the interpretation function turns into a \emph{filter} on
% the domain, letting through only those objects that are judged to be in
% the denotation of the respective predicate. We can thus locate the
% function of connecting language and world here, in the interpretation
% function of each word.\footnote{%
%   This perspective has previously been taken by \cite{schlaetal:imagewac} and developed for simple expressions; the present section builds on that work.
% }

\ \\
With this in hand, we can now explore how the available data could help us learn and evaluate lexical interpretation functions and their composition. We will look at the available expression types in order of increasing syntactic complexity. The relation that we will make use of first is the ``annotates'' relation between expressions $e$ and (parts of) images $I$. This relation gives us the value of $[\![e]\!]^{\langle D, I\rangle}$ (which is an entity or a truth value); the interest is in learning about $[\![\cdot]\!]^{\langle \cdot, I\rangle}$ as it covers the constituents of $e$ and their composition.

\subsection{Expression Types Found in Corpora}
\label{sec:extypes}

% We can keep this part short -- goal is to give overview of what's available.. 

\subsubsection{Sub-Sentential Expressions}
\label{sec:subsent}

\begin{figure}
  \begin{center}
    %\vspace*{-2.5\baselineskip}
    \includegraphics[width=0.5\columnwidth]{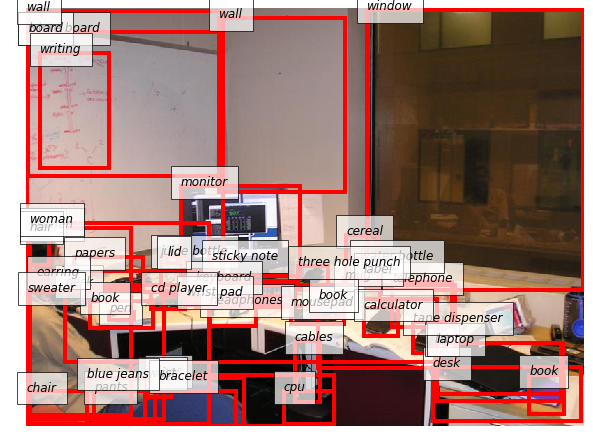}%}
    \includegraphics[width=0.5\columnwidth]{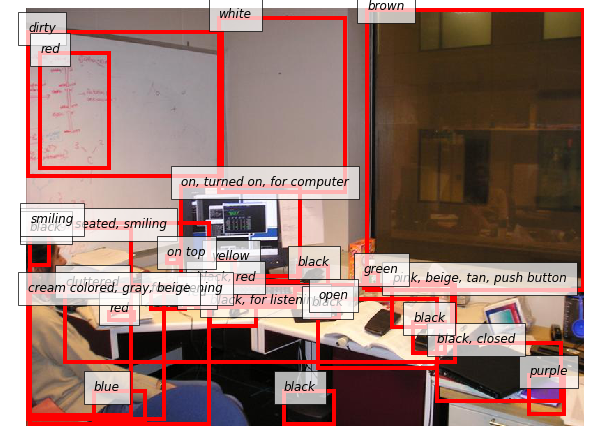}%}

  \end{center}
\vspace*{-1.6\baselineskip}
  \caption{Object ``names'' (left) and ``attributes'' (right) from \visgen, for an example image}
  \label{fig:nadj}
\end{figure}

Of the corpora discussed here, only \visgen\ provides open-class \textbf{single word} annotations.
% (\coco, for example, annotates the image objects with \emph{labels}, from a closed set; that, we argue, is a different type of annotation and less suited to learning interpretation functions.)
Objects in the corpus are associated with ``names'' (typically \textbf{nouns}), and ``attributes'' (typically \textbf{adjectives}), which were semi-automatically segmented out of larger expressions provided by annotators (to be discussed below). Figure~\ref{fig:nadj} shows this for one image from the set. (It illustrates at the same time how fine-grainedly the corpus is segmented---on average it provides 36 object bounding boxes per image.)

% \begin{figure}
%   \begin{center}
%     %\vspace*{-2.5\baselineskip}
% \end{center}
% \vspace*{-1.6\baselineskip}
%   \caption{Object ``names'' (left) and ``attributes'' (right) from \visgen, for an example image}
%   \label{fig:nadj}
% \end{figure}

The \visgen\ annotation provides over 105,000 word form types, of which about 10,500 have at least 10 instances. Using the normalisation to WordNet synsets in the corpus, this reduces to roughly 8,000 types, of which 3,500 occur at least 10 times. %As Figure~\ref{fig:ndist} shows for the names, the distribution is roughly Zipfian.
The distribution (not shown here) is roughly Zipfian---and reveals a certain bias in the data, with ``man'' occuring twice as often as ``woman'', for example.
% Depending on how one would want to set up the training (for word forms or lexemes), this
This is a sizeable vocabulary for which interpretation functions can be learned from this data.
%; the vocabulary could be further augmented with data from the referring expressions described below.

We have briefly mentioned the problem of getting \emph{negative instances} of word denotations, as required by typical machine learning methods. One method is to sample from the set of objects in a given image that are \emph{not} annotated with a word; but this requires making the aforementioned (non-warranted) exhaustivity assumption. \cite{schlaetal:imagewac} have shown  this to be unproblematic for the data that they used; establishing to what degree it would be here we leave to future work. It is likely to be more of a problem for adjectives, where the choice of what to mention is governed much more by the context than the choice of which name to use for an object.

This data can be assembled into simple nominal phrases (ADJ + N; e.g.\ ``brown window'' for top right of Figure~\ref{fig:nadj}). Semantically, these would be \textbf{indefinite noun phrases}, as all that is guaranteed is that they are appropriate for the object that they apply to (but there may be others of that type in a given image). With the denotation being known, this can be used to evaluate the semantic composition. %of the denotations of its parts.

More interesting and complex are the noun phrases found in the \textbf{referring expression} corpora (the \referit\ variants; see above). These expressions were produced in the form that they are recorded in the corpora (unlike the single word expressions discussed above), and in an actual context of use, namely with the aim to single out an object to a present interlocutor. This makes this data set also interesting from a pragmatic point of view, as one can ask how the context (in the image, but also in the production situation) may have influenced the linguistic choices.
The following shows the referring expressions available for the tennis player on the right in the image from Figure~\ref{fig:immod} above; also shown is the annotation from the \grex\ corpus:

\ex. \label{ex:refexps}
\a. \label{ex:refcoco} RefCoco: lady in black on right $|$ girl in black $|$ woman in black
\b. \label{ex:refcoco+}  RefCoco+: black shirt $|$ girl in black $|$ player in black
\b. \label{ex:grex} GoogleREX: woman in black tank top and shorts holding tennis racket $|$ woman in black outfit shaking other tennis player hand

Contrasting the \grex\ expressions illustrates the influence of the context of use on the shape of the expressions. The \grex\ annotators did not have interlocutors and were just tasked with producing expressions that describe the object uniquely. The \referit\ expressions do this as well, but additionally, they do this in the most efficient and effective way, as the players had an incentive to be as fast as possible, while ensuring referential success. This shows: The average length of \refcoco\ expressions is 3.5 token, that of \grex\ 8.3.\footnote{%
  These corpora have been used by \cite{Kazemzadeh2014}, \cite{yueatal:refcoco}, \cite{Maoetal:cocorefexp_Final}, \cite{schlaetal:imagewac}, \cite{Cirik2018b} to train and test models of
referring expression resolution.}

We also find \textbf{relational expressions} like the following in these corpora, which identify the \emph{target} object by relating it to another one (the \emph{landmark}):

\ex. \label{ex:relexp} woman under suitcase $|$ laptop above cellphone right $|$ black van in front of cab

To learn the interpretation of such relational items (here, ``under'', ``above'', ``in front of''), it would be good to have grounding information also about the landmark. The corpora mentioned so far do not give us this,\footnote{%
  For a portion of \grex, this was added by \cite{Cirik2018b}.
} and so we turn to \visgen, and away from referring expressions.

\begin{wrapfigure}[13]{r}{0.5\linewidth}
  \begin{center}
\vspace*{-1.8\baselineskip}
\includegraphics[width=0.48\columnwidth]{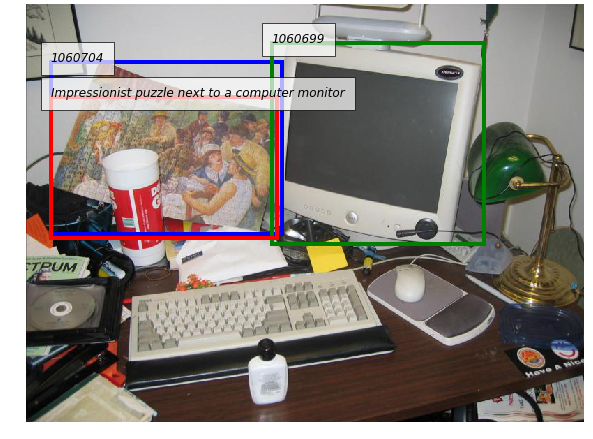}%}
\end{center}
\vspace*{-2.2\baselineskip}
  \caption{A region description from \visgen}
  \label{fig:puzzle}
\end{wrapfigure}

\visgen\ was collected with the explicit purpose of
providing material for learning ``interactions and relationships between
objects in an image'' \citep{krishnavisualgenome}. The starting point of
the annotation was the marking of a region of interest in the image, and
the annotation of that region with a ``region description'', ie.\ an
expression that is true of that region. Note the difference to referring expressions: no stipulation is made about whether it is or is not true of \emph{other} objects in the image. Annotators were encouraged to provide region descriptions that are relational, and these then form the basis of an abstracted
representation of that relation. Figure~\ref{fig:puzzle} shows an example of such a region description; the corresponding annotation is shown in \ref{ex:puzzle}, slightly re-arranged to make clearer its similarity to classical logical forms (LFs).\footnote{%
  What this also illustrates is that the normalisation decisions made in the corpus can occasionally be somewhat questionable. Here, the part ``next to a'' is normalised to the verb ``be''; presumably, the annotator added the elided copula here and rather ignored the spatial relation.
}

\ex. \label{ex:puzzle} {\small \texttt{"next to a":be.v.01(1060704:puzzle.n.01, 1060699:computer\_monitor.n.01)}}

% \ex. \label{ex:visgenreg} the clock is green in colour $|$ shade is along the street $|$ A sign on the facade of the building $|$ A tree trunk on the sidewalk $|$ A man in a red shirt $|$ The back of a white car

\begin{wrapfigure}[13]{r}{0.5\linewidth}
  \begin{center}
\vspace*{-1.8\baselineskip}
\includegraphics[width=0.47\columnwidth]{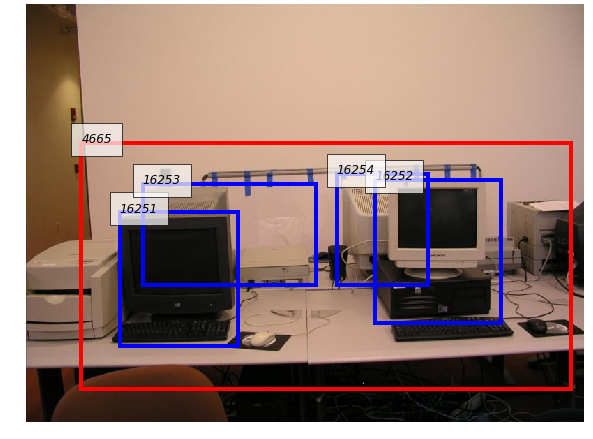}%}
\end{center}
\vspace*{-1.9\baselineskip}
  \caption{``there are desktop computers on the desk''}
  \label{fig:computers}
\end{wrapfigure}

There are over 5 million region descriptions in \visgen, of which almost 2 million are parsed into this logical form. There are around 37,000 different relational terms in this set, of which around 3,100 occur more than 10 times. From this, a sizable number of relational interpretation functions could be learned.

Before we move on, we note that in about 6.8\% of the region descriptions there is more than one object associated with an expression; as in the example in Figure~\ref{fig:computers}, where ``desktop computers'' is resolved to four different bounding boxes. Such configurations could be used to learn the function of the plural morpheme. Looking at the expressions, there are also more than 1,000 instances each of quantifiers and numerals such as ``several'', ``two'', ``many'', which provides opportunity to learn their meaning.

\subsubsection{Sentences}
\label{sec:sent}

We now turn to expressions that need to be evaluated relative to the image as a whole. Such expressions can be \emph{constructed} for example by plugging the nominal phrases from above into the sentence frame ``there is \emph{NP}$_{indef}$'' (e.g., ``there is a brown window'' for Figure~\ref{fig:nadj}), to yield \textbf{existential assertions}. Negative examples (where the constructed sentence is false) can be selected by sampling an image that is not annotated as containing an object of that type, again making use of an exhaustivity assumption.

\begin{figure}
  \begin{center}
    %\vspace*{-2.5\baselineskip}
    \includegraphics[width=0.2\columnwidth]{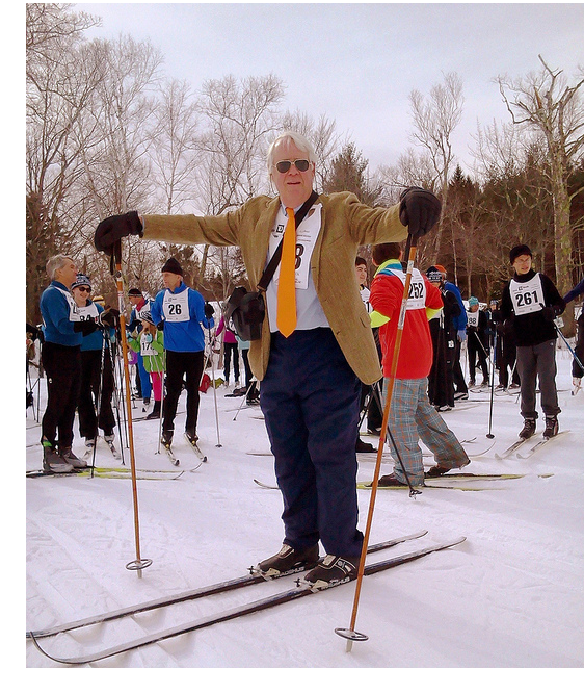}%}
    \includegraphics[width=0.3\columnwidth]{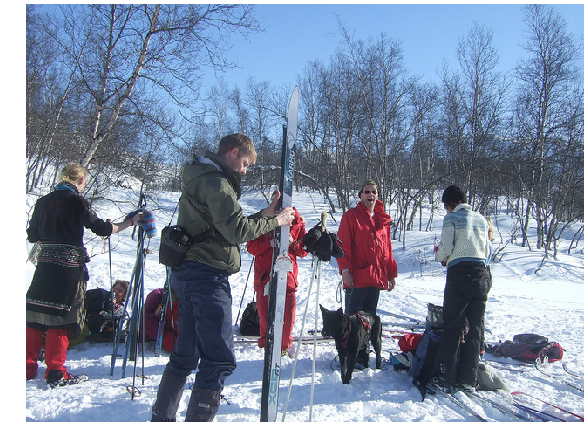}%}
    \includegraphics[width=0.3\columnwidth]{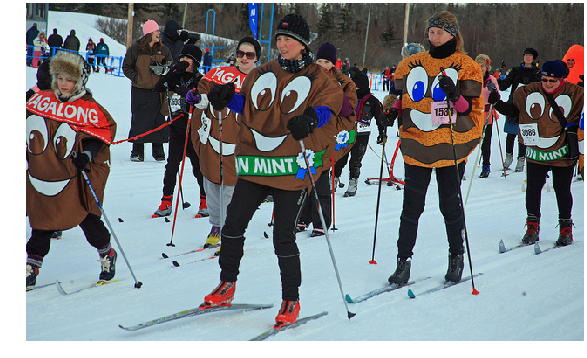}%}
  \end{center}
\vspace*{-1.6\baselineskip}
  \caption{``A man standing in the snow with skis on.'' (left), and distractors (visual similarity, middle; semantic similarity, right)}
  \label{fig:cap}
\end{figure}

Constructing examples in this way gives us control over the complexity of the expression, at the cost of a loss in naturalness. Some of the corpora, however, also come with \emph{attested} examples of expressions that are meant to describe the image as a whole; \coco\ for example provides over 400,000 of such \textbf{captions}. Figure~\ref{fig:cap} (left) gives an example of an image/caption pair.

How can we sample negative instances, where the image is \emph{not} described by the caption? One method is to simply sample an arbitrary image from the corpus: there will be a good chance that it does not fit the caption.
Too good a chance, perhaps, in that we are likely to hit an image that does not even contain any of the entity types mentioned in the expression.
To make the task harder, we can now make use of one of the derived relations described above, namely a simiarity relation between images.

We looked at two ways of defining such a relation. \emph{Visual} similarity (\(sim_{I/I}^{vis}\)) is the inverse of the cosine distance in image representation space, using a pre-trained convolutional neural network (we used VGG-19, \cite{vgg19}, pre-trained on ImageNet \cite{imagenet:2015}). We
compute \emph{content-based} or \emph{semantic} similarity
(\(sim_{I/I}^{sem}\)) by vectorising the image annotation (in a many-hot
representation with the object types as dimensions), using SVD to
project the resulting matrix into a lower-dimensional space. Given our analogy between semantic models and images, with this we then have a similarity relation between models, and we can select distractor images / models that are more challenging to refute. Figure~\ref{fig:cap} shows distractors selected via visual (middle) and semantic (right) similarity. As this illustrates, quite fine-grained resolution abilities are required to recognise these as not fitting the caption. (Man, but skis not on; man with skis on, but not standing.)

Captions from \coco\ are not finely grounded (no links between objects in the image and parts of the expression). \flickr\ provides this; for reasons of space, we do not show an example here. We also skip over the \emph{wh}-questions (with answers) that are available for \coco\ and \visgen, noting only that they add an interesting generation challenge (if set up as open answer task; if set up as multiple-choice task, this reduces to making a decision for a proposition).

\subsubsection{Discourses}
\label{sec:disc}

\begin{figure}
  \begin{center}
    %\vspace*{-2.5\baselineskip}
    \includegraphics[width=0.35\columnwidth]{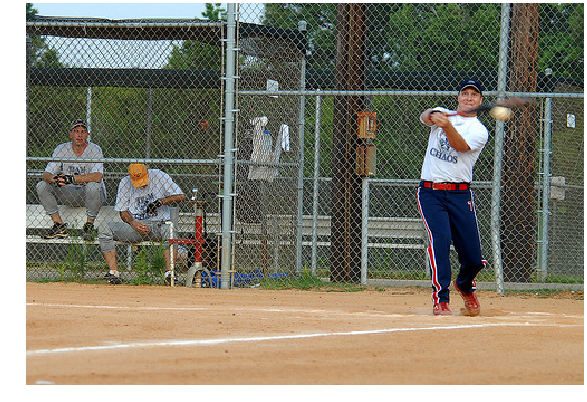}%}
    \includegraphics[width=0.28\columnwidth]{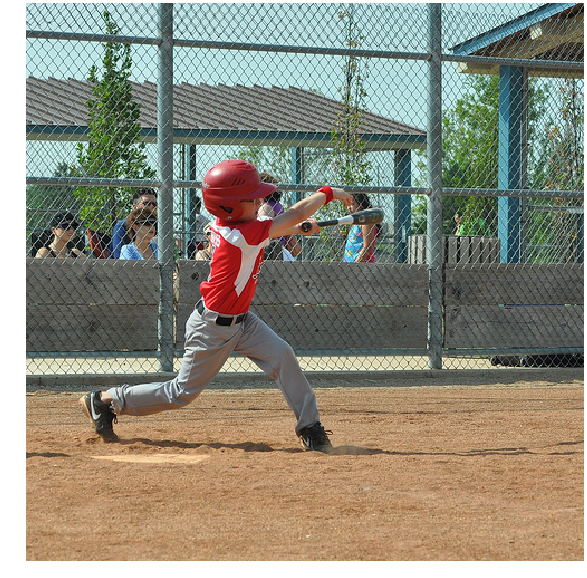}%}
    \includegraphics[width=0.18\columnwidth]{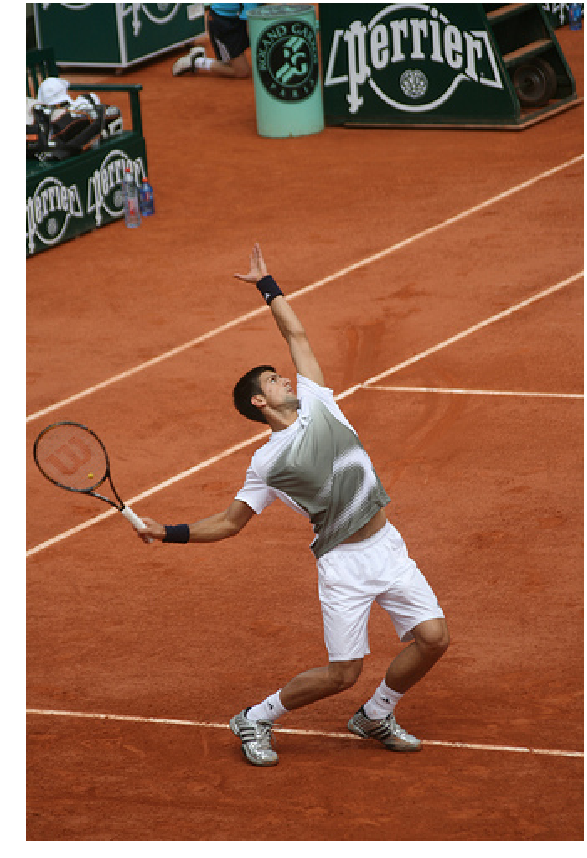}%}
  \end{center}
\vspace*{-1.6\baselineskip}
  \caption{Image described by paragraph (see text; left), and distractors (visual similarity, middle; semantic similarity, right)}
\vspace*{-1\baselineskip}
  \label{fig:par}
\end{figure}

Finally, example~\ref{ex:par} shows an \textbf{image description paragraph} for a \visgen\ image. The associated image and two distractors are shown in Figure~\ref{fig:par}. The semantic challenge here when evaluating such a paragraph relative to an image, at least when a probabilistic approach is taken, is that a decision must be made on how to combine the uncertain judgements from each constituent sentence.

\ex. \label{ex:par} The baseball player is swinging the bat. The ball is in the air. The dirt on the
ground is light brown. The baseball player is wearing blue pants. The other
baseball players are watching from The Dugout. The baseball player swinging the
bat is wearing a dark-colored baseball hat. He's also wearing a bright red belt.

%For reasons of space, we will only look at \emph{constructed} discourses (which have the advantage of their parts being fully grounded) below for expression / expression tasks and only note here that they can be constructed out of sequences of \visgen\ region descriptions.
%\das{End with table that summarises the data sets?}

As this survey has shown, there is plenty of data available for learning grounded interpretation functions for individual words (nouns, adjectives, prepositions), and for evaluating (or even learning) how these functions must be put together to yield interpretations for larger expressions (NPs, sentences, and even discourses).

\section{Expressions and Implications}
\label{sec:ent}

\subsection{Images as Implicit Link between Expressions}
\label{sec:link}

Besides the question of whether a statement is true of a given
situation, an interesting question often is whether a statement
\emph{follows} from another one. There are various ways of tying down
what exactly ``follows'' may mean. A very general one is
given by \cite{chierchi:meaning}, who use ``A implies B'' for cases
where (the statement and acceptance of) A \emph{provides reason} to also
accept B.\footnote{%
  This is also how later the influential ``recognising textual entailment'' challenge \citep{Dagan:rte} would describe the relation, however also starting the tradition in natural language processing to overload the term ``entailment'' to cover all of what could more generally be called ``implication''. \cite{youngetal:flickr30k} call their task, defined via images as well and our inspiration for the work described here, with a qualifier as ``\emph{approximate} entailment''.}
This covers cases where a \emph{proof} can be given that
connects B to A (where the relation would be \emph{syntactic
consequence}, \(\vdash\)), cases where an argument can be made that any
model that makes A true will also make B true (\emph{semantic
consequence}, \(\models\)), but also cases where A may just make B very
plausible, given common sense knowledge (which we might call
\emph{common sense implicature}, and denote with \(\models_{cs}\)).

% By now, the recipe should be clear.
Here, we look at relations that we can take from the corpora and ask whether these can help us get at these semantic implicature relations. We make use
of the fact that for most of the image objects in the corpora, we have available more than one expression of the same type, e.g., more than one referring expression, or more than one caption.
% Can we make use of this \emph{annotates-same-image} relation?
In the following, we take a look at some examples, sorting the discussion by the type of expressions that we pair.\footnote{
  Our inspiration for this approach comes from two sources. As mentioned, \cite{youngetal:flickr30k} used image captions to create their ``approximate entailment'' data sets; our proposals here can be seen as a generalisation of this to other pairings. Further, the original ``natural language inference'' dataset by \cite{snli:emnlp2015} used captions as seeds, but had the entailments and contradictions manually generated and not derived via image relations, as we do here.
}
We will argue that to predict the presence (or not) of an implicature relation, a different, complementary kind of lexical knowledge is required than for evaluation relative to an image (or situation); cf.\ \cite{marconi:lexcomp}.

\subsection{Types of Relations}
\label{sec:rels}

\subsubsection{Same Level / Rephrasing}
\label{sec:rephr}

Example \ref{ex:refexps} shows referring expressions from \refcoco\ (left) paired with another expression referring to the same object (middle) and with one referring to a randomly sampled other object from the corpus. The prediction task is to identify the left/middle pair as standing in an implicature relation, and the left/right pair as not standing in this relation. (To put a practical spin on it, this could be seen as detecting whether the second pair part could be a \emph{reformulation} of the first, perhaps as response to a clarification request.)

\ex. \label{ex:refexps}
\a. \label{ex:refa} right girl on floor $||$ lady sitting on right $|$ guy on right
\b. \label{ex:refb} woman $||$ left person $|$ pizza on bottom right
\c. \label{ex:refc} man trying to help with suitcase $||$ man in jacket $|$ very top zebra

Despite the brevity of the expressions, as this example indicates, this task seems to require quite detailed lexical knowledge, for example detecting incompatiblity between ``guy'' and ``girl'' in \ref{ex:refa}, but compatiblity between ``lady'' and ``girl''. (If this knowledge were available, perhaps a \emph{natural logic}-type \citep{moss:naturallogic} approach could then be taken.)
Creating this dataset only requires that several referring expressions are available for the same object, and indeed \refcoco\ for example provides on average 7.1 per object, for a total of over 140,000 referring expressions.

We randomly sampled 60 instances of such pairings (balanced pos/neg) and presented each to three workers on Amazon Mechanical Turk, asking for a semantic relatedness judgement (on a 4 step Likert-scale). Using the majority label and binning at the middle of the scale, the accuracy is 0.68. This indicates that while noisy, this method creates a recognisable semantic relation between these expressions.\footnote{%
  Note that the task was to judge pairs, not to decide between two hypotheses, which would presumably be a simpler task.
}

Example \ref{ex:capsent} shows similar pairings of captions, with the negative instance (the final part of each sub-example) taken from a distractor image selected for semantic image similarity. As this illustrates, the task only becomes harder, with the caption that is intended to be non-matching occasionally accidentally even intuitively being compatible after all. (Crowd accuracy, henceforth AMT, with same setup as described above: 0.63.)
\ex. \label{ex:capsent}
\a. \label{ex:capsenta}
A woman with a painted face riding a skateboard indoors. $||$
A woman with face paint on standing on a skateboard. $|$
There are men who are skateboarding down the trail.
\b. \label{ex:capsentb}
Man and woman standing while others are seated looking at a monitor. $||$
A man and woman play a video game while others watch. $|$
Two people standing in a living room with Wii remotes in their hands.

\subsubsection{More Specific / Entailment}
\label{sec:ent}

Since some of the corpora overlap in their base image data, we can intersect the annotation and create derived data sets. \ref{ex:capob} shows examples of a caption from \coco\ (left) paired with an object from \visgen\ (slotted into a ``there is (a) \_\_'' frame for presentation) taken from the same image (middle), and a randomly sampled object (right) in \ref{ex:capoba} and \ref{ex:capobb}, and with region descriptions (also from \visgen) in the other examples. 

\ex. \label{ex:capob}
\a. \label{ex:capoba}
A man wearing a black cap leaning against a fence getting ready to play baseball. $||$ there is (a) man $|$ there is (a) cow
\b. \label{ex:capobb}
Rice, broccoli, and other food items sitting beside each other
$||$ there is (a) health foods $|$ there is (a) granite
\c. \label{ex:caprega}
A man playing Wii in a room $||$
there is/are (a) a plant that sits on a desk $|$
there is/are (a) field covered in green grass
\d. \label{ex:capregb}
A woman is riding a wave on a surfboard. $||$
there is/are (a) Woman with the surfboard. $|$
there is/are (a) Students sitting at their desks

Judging from these examples, quite detailed knowledge about situations and possible participants seems to be required to predict these relations. (AMT accuracy caption/object: 0.58, caption/region: 0.6.)

\subsubsection{More Detailed / Elaboration}
\label{sec:elab}

Finally, \ref{ex:cappar} shows examples of a caption (from \coco) paired with a paragraph (from \visgen -paragraphs) describing the same (middle) or another, but similar image. The task here is to detect whether the extended description fits with the short description or not, which again seems to require quite detailed knowledge about situations and likely sub-events. (AMT: 0.6.)

\ex. \label{ex:cappar}
%\a. \label{ex:cappara}
two people lying in a bunk bed in a bedroom.\\ %$||$
A boy and girl are sitting on bunk beds in a room. The boy is wearing a red shirt and dark pants. The girl is wearing a gray shirt and blue jean pants. There is a green and pink blanket behind the boy on the top bunk. The girl is sitting on a rolled up blanket. She is wearing red glasses on her eyes.
\\%$|$
A woman is sitting on a bed beside a little girl. She is wearing a sweater and black bottoms. The woman has eyeglasses on her eyes. The girl is wearing a colorful jacket. The girl is looking at a book that is opened on her lap. The bed is sitting against a white painted wall. There is a red blanket on the bed.

\vspace*{.3\baselineskip}
Using this general recipe, further datasets can be created with other combinations, for example pairing sets of region descriptions with further descriptions either from the same or from a different scene, or for the task of predicting the number of distinct entities introduced by a sequence of region descriptions. For reasons of space, we do not show examples here.

\section{A Case Study: Model-Building for Predicting Entailment}
\label{sec:modbuild}

\begin{wrapfigure}[13]{r}{0.5\linewidth}
  \begin{center}
\vspace*{-2\baselineskip}
\includegraphics[width=0.5\columnwidth]{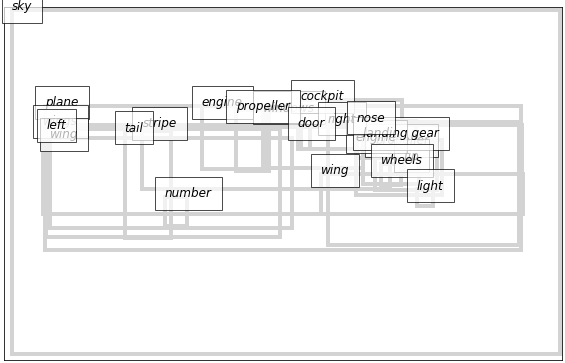}%}
\end{center}
\vspace*{-1.6\baselineskip}
  \caption{A Retrieved Abstract Exemplar Situation}
  \label{fig:retrieved}
\end{wrapfigure}

Entailment tasks, triggered by the aforementioned ``natural language inference'' dataset \citep{snli:emnlp2015} have in recent years become a staple NLP task. They are typically tackled with very high-capacity machine learning models that classify distributed representations of the candidate relata, e.g., as in \citep{Devlin2018}. With the perspective developed here, we can liken such approaches to the \emph{syntactic} way of defining entailment ($\vdash$), in that these approaches only take the surface form into account (and implicitly learn and use the required common sense knowledge).

A \emph{semantic} approach seems possible as well, however. In its brute force form, it would implement the typical way in which \emph{semantic consequence} is defined, by quantification over all models. Here, this would mean testing, along the lines developed in Section~\ref{sec:ei}, whether all images (in a sub-corpus held for that purpose) that make the premise true also make the hypothesis true. We try something else here, which is more like model-building \citep{Bos2003}, for data of the type illustrated in \ref{ex:capob} above.

The idea is as follows. Given the premise (in our case, always a caption), we retrieve a set of images (other than that from which the caption was taken), via captions that are nearest neighbours in a text embedding space (for which we used the ``universal sentence encoder'' by \cite{Cer2018}). That is, we make use of a derived expression/expression relation, to create a relation between an expression and a set of retrieved models. (One can think of these as situation exemplars stored in memory and retrieved via their short descriptions.) Figure~\ref{fig:retrieved} shows such a retrieved model (abstracted away from the actual image content, which is not used), for the trigger caption ``An airplane flying through the sky on a cloudy day.'' and retrieved via its most similar caption ``White and blue airplane flying in a grey sky.''.

We then test the candidate expressions (or rather, their ``logical forms'', as given by \visgen) against this set of models. For objects (as in \ref{ex:capoba} and \ref{ex:capobb} above), this checks whether an object of the appropriate type is in the retrieved models; for relations, this additionally checks whether the relation is also present. If the required types are present in all models, this would yield a score of 1. We set a threshold (in our experiments, at 0.2), above which a positive decision is made. As baseline, we use token overlap between premise and hypothesis for objects and intersection over union for the longer region descriptions, and distance in the embedding space. We created 10,000 triples each for the caption/object and the caption/region task. 

The results in Table~\ref{tab:results} indicate that this rather simple model captures cases that the baselines do not. An example where this is the case is shown in \ref{ex:modbuild}; here the retrieved models seem to have provided the entities (``umpire'' and ``jacket'') which are likely to be present in a baseball scene, but aren't literally mentioned in the premise.

% \begin{table}[h]
%   \centering
%   \begin{tabular}[h]{ccc}
%     & Model & Baseline\\
%     \hline
%     Captions / Objects & 0.67 & 0.57\\
%     Captions / Regions & 0.66 & 0.55
%   \end{tabular}
%   \caption{Results}
%   \label{tab:results}
% \end{table}

\begin{table}[h]
  \centering
  {\small
  \begin{tabular}[h]{cccc||cccc}
    Task & Model & Strg.Bsln & Embd.Bsln & Task & Model & Strg.Bsln & Embd.Bsln\\
    \hline
    Captions / Objects & 0.67 & 0.58 & 0.64 & Captions / Regions & 0.65 & 0.54 & 0.50
  \end{tabular}
  }
  \vspace*{-.4cm}
  \caption{Results for Predicting Entailment via model retrieval (and baselines)}
  \label{tab:results}
\end{table}

\ex. \label{ex:modbuild}
Baseball batter hitting ball while other players prepare to try and catch it. $||$ jacket worn by umpire $|$ silverware on a napkin

This is clearly not more than a first proof-of-concept. We've included it here to motivate our tentative conclusion that the perspective introduced in this paper might have value not only for deriving interesting data sets, but also for tackling some of the tasks. In future work, we will explore methods that directly predict image layouts [e.g., \citep{Tan2018}], comparing them to direct prediction approaches and evaluating whether the former methods offer a plus in interpretability through the step of predicting abstract models.

\section{Conclusions}
\label{sec:conc}

Our goal with this paper was to show, with detailed examples and descriptive statistics, that language / vision corpora can be a fertile hunting ground for semanticists interested in grounded lexical semantics. There is data pairing various, ever more complex, kinds of expressions with image objects (either parts of images, or images as a whole). Moreover, using these corpora, data sets can be derived that pair expressions, where a semantic relation holds between the parts that is recognisable to naive annotators (if not alway very clearly). As an example, we've used the perspective of treating images as models to retrieve exemplar models via language descriptions (captions), and probe those for the likely presence of entities and relations in a mentioned situation. It is our hope that this perspective might be useful to other researchers, and with the code released with this paper, we invite everyone to ask their own questions of the data, and to implement ideas on how to learn grounded interpretation.

% suggest a particular way how evaluation of expressions relative to images can be set up (namely, by augmenting the traditional compositional semantics method only slightly, letting the interpretation functions take images as input).

\paragraph{Acknowledgements} This work was done while I was at Bielefeld University and supported by the Cluster of Excellence Cognitive Interaction Technology ``CITEC'' (EXC 277), which is funded by the German Research Foundation (DFG). I thank Sina Zarrie{\ss} and the anonymous reviewers for comments.

%\clearpage
\bibliographystyle{chicago}
\bibliography{/Users/das/work/projects/MyDocuments/BibTeX/all-lit.bib}

\end{document}